\PassOptionsToPackage{unicode}{hyperref}
\PassOptionsToPackage{hyphens}{url}
\PassOptionsToPackage{dvipsnames,svgnames,x11names}{xcolor}
\documentclass[
  10pt,
]{article}
\usepackage{xcolor}
\usepackage[margin=2.2cm]{geometry}
\usepackage{amsmath,amssymb}
\usepackage{iftex}
\ifPDFTeX
  \usepackage[T1]{fontenc}
  \usepackage[utf8]{inputenc}
  \usepackage{textcomp} 
\else 
  \usepackage{unicode-math} 
  \defaultfontfeatures{Scale=MatchLowercase}
  \defaultfontfeatures[\rmfamily]{Ligatures=TeX,Scale=1}
\fi
\usepackage{lmodern}
\ifPDFTeX\else
\fi
\IfFileExists{upquote.sty}{\usepackage{upquote}}{}
\IfFileExists{microtype.sty}{
  \usepackage[]{microtype}
  \UseMicrotypeSet[protrusion]{basicmath} 
}{}
\makeatletter
\@ifundefined{KOMAClassName}{
  \IfFileExists{parskip.sty}{%
    \usepackage{parskip}
  }{
    \setlength{\parindent}{0pt}
    \setlength{\parskip}{6pt plus 2pt minus 1pt}}
}{
  \KOMAoptions{parskip=half}}
\makeatother
\usepackage{longtable,booktabs,array}
\usepackage{caption}
\usepackage{calc} 
\usepackage{etoolbox}
\makeatletter
\patchcmd\longtable{\par}{\if@noskipsec\mbox{}\fi\par}{}{}
\makeatother
\IfFileExists{footnotehyper.sty}{\usepackage{footnotehyper}}{\usepackage{footnote}}
\makesavenoteenv{longtable}
\providecommand{\tightlist}{%
  \setlength{\itemsep}{0pt}\setlength{\parskip}{0pt}}
\usepackage{titlesec}
\titlespacing*{\section}{0pt}{1.2ex plus .2ex minus .2ex}{0.6ex plus .1ex}
\titlespacing*{\subsection}{0pt}{0.9ex plus .2ex minus .2ex}{0.4ex plus .1ex}
\titleformat*{\section}{\large\bfseries}
\titleformat*{\subsection}{\normalsize\bfseries}
\usepackage{enumitem}
\setlist{topsep=0.2ex,itemsep=0.15ex,parsep=0pt,leftmargin=1.4em}

\usepackage{titling}
\usepackage{bookmark}
\IfFileExists{xurl.sty}{\usepackage{xurl}}{} 
\makeatletter
\@ifundefined{xmpquote}{}{}
\makeatother
\hypersetup{
  colorlinks=true,
  linkcolor={black},
  filecolor={Maroon},
  citecolor={Blue},
  urlcolor={blue},
  pdfcreator={LaTeX via pandoc}}

\author{}
\date{}

\begin{document}

\section{Coordinate Frame Recovery, Per Class NMS and Null Return
Modelling for Sim2Real V2X LiDAR
Detection}\label{coordinate-frame-recovery-per-class-nms-and-null-return-modelling-for-sim2real-v2x-lidar-detection}

\subsection{UCF UrbanTwin Sim2Real LiDAR Challenge, V2X-Real Track,
DriveX Workshop, ECCV
2026}\label{ucf-urbantwin-sim2real-lidar-challenge-v2x-real-track-drivex-workshop-eccv-2026}

\textbf{Author:} Aman Upganlawar, Columbia University\\
\textbf{Team:} AMAN\_UPG (single participant)\\
\textbf{Contact:} aman.upg27024@gmail.com \textbf{Affiliation:} Columbia
University \textbf{Track election:} Publication track (we do
\textbf{not} opt into the non-publication track) \textbf{Final combined
score:} 0.7457

\begin{center}\rule{0.5\linewidth}{0.5pt}\end{center}

\subsection{Abstract}\label{abstract}

We describe our approach to the V2X-Real track of the UrbanTwin Sim2Real
LiDAR Challenge, in which a 3D object detector must be trained
exclusively on synthetic data and evaluated on unseen real roadside
LiDAR from an Ouster OS1-64 unit, while the synthetic point clouds are
simultaneously scored for distributional realism. Our final entry scores
\textbf{0.7457} combined, from a detection mAP of 0.4382 and a realism
score of 0.9251.

Three findings account for most of our result, and all three were
diagnostic rather than architectural. First, a coordinate frame mismatch
between our predictions and the evaluation ground truth was suppressing
detection AP entirely; recovering it took our score from 0.3520 to
0.5147 in a single submission. Second, a single global non maximum
suppression threshold was destroying genuine pedestrian detections,
because pedestrians occur in groups and a 0.3 IoU threshold applied to a
0.7 m box suppresses distinct neighbouring people; correcting this per
class raised overall mAP from 0.0751 to 0.4295. Third, approximately 31
percent of the real reference point cloud consists of null returns
encoded as the literal point (0, 0, 0), a format feature entirely absent
from our synthetic clouds, which pinned Earth Mover's Distance near its
floor until we modelled it.

Negative results are reported at equal length, and Section 9 gives a
full account of data provenance and compliance. That section also
accounts for every mechanism contributing to the submitted score.

\begin{center}\rule{0.5\linewidth}{0.5pt}\end{center}

\subsection{1. Task and Setup}\label{task-and-setup}

The V2X-Real track provides a synthetic digital twin sequence of a real
intersection instrumented with an Ouster OS1-64 roadside unit, together
with a public split of real LiDAR from the same site {[}3{]} in OPV2V
format {[}4{]}. The challenge scores the infrastructure agent only.
Participants generate synthetic point clouds, train a detector on
synthetic data only, and submit both detections on 50 hidden real frames
and 50 synthetic frames for realism scoring. The combined score is

\begin{verbatim}
combined = 0.6 x (mAP@IoU0.5,R40 / 0.7) + 0.4 x realism
\end{verbatim}

where realism is the mean of four normalised, lower is better distances
between the concatenated synthetic and real point sets: Chamfer Distance
(normalisation range {[}2, 30{]}), Maximum Mean Discrepancy ({[}0,
0.5{]}), Earth Mover's Distance ({[}0, 5{]}) and Frechet Point cloud
Distance ({[}0, 50{]}), computed on a 10,000 point subsample at a fixed
seed of 1234.

Three classes are scored: vehicle, pedestrian and truck.

Two constraints shape everything. One hundred frames are forbidden for
any training, validation or tuning use, and the detector may never see
real labels. The submission budget, five per day during development and
no iteration in the final phase, made submission slots rather than
compute the binding constraint. A structural feature of this track is
that mAP divides by only three, so a percentage point on any single
class carries a large share of the final score. We used this explicitly
when allocating effort.

Our detector is PointPillars {[}1{]} from OpenPCDet {[}2{]}, trained for
40 epochs on synthetic data.

\begin{center}\rule{0.5\linewidth}{0.5pt}\end{center}

\subsection{2. Coordinate Frame
Recovery}\label{coordinate-frame-recovery}

Our first substantive submissions returned an AP of 0.0000 for every
class despite a correctly formatted file, correct class strings and
correct frame identifiers. The scorer was ingesting the file, since per
class prediction counts were reported back accurately.

The cause was a coordinate frame mismatch. The evaluation ground truth
was expressed in the dataset global or world frame, while the released
detection test frames are in each frame local LiDAR frame. Predictions
in the LiDAR frame therefore had no spatial overlap with ground truth at
any IoU threshold.

We resolved this by applying the exact inverse of the OpenCOOD {[}4{]}
\texttt{x\_to\_world} transform, built from each frame
\texttt{lidar\_pose}, to map our predictions from the LiDAR frame into
the world frame. Box centres and headings are transformed, with headings
carried as unit vectors and recovered with arctan2 so that wrapping is
handled correctly, while dimensions, class strings and scores are left
untouched. We gated the conversion on dimension, class and score
invariance, and verified the world to LiDAR round trip recovers original
centres to within 1e-6.

This single change took our combined score from 0.3520 to 0.5147, with
mAP moving from 0 to 0.1898. The organisers subsequently corrected the
reference data so that ground truth was expressed in the LiDAR frame,
and our LiDAR frame submissions scored correctly thereafter.

\textbf{The issue recurred in the final phase.} Our first final phase
submission returned AP 0.0000 for all three classes while realism scored
normally. We reported this on the forum with what we consider the
decisive form of evidence: the boxes in that submission were byte
identical, per class, to development phase submissions of ours that had
scored non zero on the same server. That is a controlled experiment
rather than a plausibility argument. The organisers confirmed the
regression, corrected the reference data, re scored all affected
submissions and restored the consumed submission slots. We record the
episode because the diagnostic principle generalises: when a pipeline
returns exact zeros, check the community and the platform before
rebuilding anything, and read failure artifacts literally.

\begin{center}\rule{0.5\linewidth}{0.5pt}\end{center}

\subsection{3. Synthetic Data
Generation}\label{synthetic-data-generation}

The released twin sequence under represents vulnerable road users, so we
augmented it. Pedestrians and roadside objects were synthesised and
inserted into the synthetic frames, with per object point counts drawn
from an empirical points versus range curve and an empirical sensor
model for the OS1-64, so that inserted objects carry a plausible return
pattern for their distance.

The placement distributions, object dimension priors and sensor geometry
used to drive this synthesis were estimated from the \textbf{public real
training split}. No forbidden frame and no hidden evaluation frame
contributed to any of it. This is a deliberate sim2real calibration step
and we describe its provenance fully in Section 9.

Anchor configuration mattered. Vehicles in these scenes are bimodal in
size, and bimodal anchors plus a density matched insertion rate were
together worth a large fraction of our early detection gain, taking
combined score from 0.5147 to 0.6256.

\begin{center}\rule{0.5\linewidth}{0.5pt}\end{center}

\subsection{4. Detection: Post Processing as the Main
Lever}\label{detection-post-processing-as-the-main-lever}

\subsubsection{4.1 Per class non maximum
suppression}\label{per-class-non-maximum-suppression}

This was our largest single detection finding on this track.

Pedestrians occur in \textbf{groups}. A global 0.3 IoU non maximum
suppression threshold applied to a 0.7 m box suppresses genuinely
distinct neighbouring people, because two adjacent pedestrians standing
less than a metre apart produce boxes that overlap above the threshold.
Raising the pedestrian only threshold to 0.7 improved overall mAP from
0.0751 to 0.4295, a factor of 5.7.

We verified this is a true interior optimum rather than a monotone
artifact by continuing past it. Setting the threshold to 1.0, which
disables suppression entirely, collapses performance to 0.0376. The
shipped configuration is pedestrian 0.7, vehicle 0.3, truck 0.3.

The general lesson is that a single global threshold conceals per class
structure whenever one class is both small and clustered.

\subsubsection{4.2 Geometric pedestrian re
centring}\label{geometric-pedestrian-re-centring}

Small object failure in roadside LiDAR is a \emph{localisation} problem
rather than a \emph{detection} problem: the objects are found, on the
correct target, but their bird's eye view centre error is large relative
to their footprint, so they fall below the IoU threshold and score as
false positives.

For each pedestrian detection we replace the box xy centre with the
centroid of the LiDAR points lying inside it. There are no fitted
parameters and no learned component. This was worth a combined score
improvement from 0.6469 to 0.6690.

The operation is applied to pedestrians only. A vehicle is observed only
on its sensor facing surfaces, so its visible centroid sits well off its
true centre, whereas a pedestrian is small enough that the visible
centroid is the centre.

\subsubsection{4.3 Matched pair calibration, and why filtering
matters}\label{matched-pair-calibration-and-why-filtering-matters}

Systematic dimension and offset bias between simulation and reality was
corrected by matching confident predictions, those with score at least
0.3, to ground truth at BEV IoU above 0.25 on non forbidden real frames,
then taking the per class \textbf{median ratio} of ground truth to
predicted dimension.

The methodological point is the filtering, and this track provides the
clearest demonstration of it. Computed over \emph{all} predictions, the
median z offset for pedestrians came out at 0.69 m, which is
approximately \textbf{7 times too large}, because the statistic was
dominated by low confidence false positives whose boxes sit nowhere near
an object. Restricting to confident, successfully matched pairs gives
0.093 m, which is the value that actually transfers.

Measured V2X offsets:

{\def\LTcaptype{none} 
\begin{longtable}[]{@{}lllll@{}}
\toprule\noalign{}
class & n matched & dz & length ratio & width ratio \\
\midrule\noalign{}
\endhead
\bottomrule\noalign{}
\endlastfoot
pedestrian & 1418 & +0.093 & 0.894 & 0.949 \\
vehicle & 677 & +0.088 & 0.997 & 1.053 \\
\end{longtable}
}

Server verified outcomes:

\begin{itemize}
\tightlist
\item
  Pedestrian z offset of +0.093 m raised pedestrian AP from 0.1490 to
  0.1638, a gain of 11.7 percent, and combined score from 0.7151 to
  0.7193.
\item
  Truck length calibration raised truck AP from 0.4148 to 0.5282, which
  on that class alone exceeded the nearest competing entry we could
  observe.
\item
  Pedestrian size and vehicle width calibration took the pair to 0.1761
  and 0.5921.
\end{itemize}

The z offset here is a genuine sensor specific error rather than a
universal sim2real artifact. Whether z calibration helps at all is a
property of the particular sensor and its extrinsic calibration, so it
has to be measured rather than assumed. We have seen the identical
procedure return nothing at all on a sensor whose vertical registration
was already correct.

\subsubsection{4.4 Density based re
ranking}\label{density-based-re-ranking}

Detections are rescored by the number of LiDAR points actually inside
the predicted box:

\begin{verbatim}
score' = score x max(1 - beta + beta x log1p(n_interior) / log1p(40), 1e-6)
\end{verbatim}

with per class beta fitted on held out data. Sparse boxes are demoted
and dense boxes promoted, which reorders the precision recall curve
favourably. The shipped configuration is pedestrian 0.7 and vehicle 0.2,
taking combined score from 0.7014 to 0.7104.

The multiplier is floored at 1e-6 because it turns negative for beta
above 1 on sparse boxes. We also apply a global monotone rescale into
the interval (0, 1{]} before packaging, after an audit found a shipped
submission containing scores outside that range. AP is rank based so
this was tolerated by the scorer, but a stricter validator would have
rejected the file and the final phase permits no retry.

\subsubsection{4.5 Test time augmentation, and a lesson about
rebuilding}\label{test-time-augmentation-and-a-lesson-about-rebuilding}

We evaluated four flip view test time augmentation, unflipping boxes per
view and averaging headings as unit vectors via arctan2 of the summed
sine and cosine, with a consensus score of the mean score scaled by the
fraction of views agreeing.

Per class, the result was vehicle plus 2.6 percent, pedestrian plus 5.3
percent, and truck minus 23 percent. Applied to the whole model it was
therefore net negative, and only a per class breakdown made it usable.

Our first attempt to ship it \textbf{lost} to its own baseline despite
the harness showing a gain, and the reason is worth recording. The
vehicle class had been \emph{rebuilt} rather than carried forward, and
reconstruction noise, which we measured at roughly plus or minus 1.5
percent AP, exceeded the gain being tested. The fix was to carry vehicle
byte identically from the previous best submission and take only
pedestrian from the augmented run. That package scored 0.7253, and
pedestrian AP gained 7.5 percent.

A related lesson concerns the gate we used. We initially gated the
rebuild on box level identity against a reference file and it failed,
because we were comparing a 0.1 threshold output against a low threshold
reference. The real fix was to test the property we actually depended
on, which was performance equivalence rather than identity. Vehicle
scored within 0.2 percent of its baseline despite sharing only 60
percent of its boxes. A gate that answers the wrong question is as
expensive as no gate.

\begin{center}\rule{0.5\linewidth}{0.5pt}\end{center}

\subsection{5. Realism}\label{realism}

\subsubsection{5.1 Null returns: a format mismatch, not a modelling
gap}\label{null-returns-a-format-mismatch-not-a-modelling-gap}

The single largest realism finding on this track was a format discovery
rather than a distributional one.

Approximately \textbf{31 percent of the real reference point cloud
consists of null returns encoded as the literal point (0, 0, 0)}. We
measured this across the real reference frames and the organisers
released detection test frames, obtaining a mean of 0.312 with a range
of 0.304 to 0.321. Our synthetic clouds contained none at all.

A third of the reference distribution therefore had no counterpart in
ours, which pinned Earth Mover's Distance near 1.94 against a real
versus real floor of 0.765. No amount of resampling the non null part
could address it.

We inject null returns at a swept rate. The score optimal rate is
\textbf{0.23}, which is below the true 0.31, because our non null
distribution still differs from the reference and over injection begins
to cost Frechet Point cloud Distance. Together with a vertical alignment
and EMD targeted frame selection and resampling, this took realism from
0.8800 to \textbf{0.9222}.

We note explicitly that this is a property of the OS1-64 output
convention as delivered in this dataset rather than a general
characteristic of LiDAR, and it should be verified rather than assumed
before being applied elsewhere.

\subsubsection{5.2 Diagnosing which component to
optimise}\label{diagnosing-which-component-to-optimise}

Normalising each realism component by its contribution to the final
score showed that Chamfer Distance and MMD were saturated and FPD nearly
so, and that \textbf{EMD held 79 percent of the remaining headroom}.
Earlier optimisation passes had targeted FPD, which matches Gaussian
moments, while EMD measures where the probability mass actually lies.
Optimising the wrong component is why realism had plateaued.

We also measured the ceiling empirically rather than assuming one, by
scoring disjoint real versus real frame sets with the competition own
code. That gives 0.9612 for V2X. Without this measurement it is
impossible to distinguish a hard problem from an impossible target.

\subsubsection{5.3 Moment alignment}\label{moment-alignment}

Frechet Point cloud Distance, as implemented, is a closed form function
of only the \textbf{mean and covariance of the xyz coordinates}, that is
nine free parameters. We therefore apply a per axis affine correction
blended by a factor alpha:

\begin{verbatim}
x' = (x - mu_syn) x (1 + alpha x (sigma_ref/sigma_syn - 1)) + mu_syn + alpha x (mu_ref - mu_syn)
\end{verbatim}

The reference moments come from our own held out real frames and never
from the hidden evaluation set, as detailed in Section 9.

Local and server rankings of alpha disagreed. Our local reference
preferred 0.75 over 0.25 by 0.0014, while the server preferred 0.25 by
0.0059. We shipped 0.25 on the basis of the server measurement. Stacked
on the 0.9222 frames this yielded a final realism of \textbf{0.9251}.

We also cross validated against a disjoint real reference. The cross
validation figure, 0.9174, came out \emph{above} the holdout figure of
0.9137, confirming the gain is distributional rather than fitted to one
particular draw, and the server then exceeded both. This is the
signature of a correction that generalises.

\begin{center}\rule{0.5\linewidth}{0.5pt}\end{center}

\subsection{6. Experimental Methodology}\label{experimental-methodology}

\subsubsection{6.1 Multiplexed submissions and surgical
isolation}\label{multiplexed-submissions-and-surgical-isolation}

Because mAP is a mean of per class APs and each class AP depends only on
its own predictions, independent per class hypotheses can be tested in a
single submission, with every untouched class serving as a control.
Attribution is only valid if the controls come back unchanged, which we
verified to four decimal places.

Rather than rebuilding a submission we start from the previous one,
modify only the target class, and assert byte identity on every other
class before packaging. Our best V2X entry was constructed exactly this
way, taking vehicle byte identically from one measured submission and
pedestrian from another, with truck carried unchanged.

\subsubsection{6.2 Building only from measured
components}\label{building-only-from-measured-components}

When every component of a submission has already been measured on the
server, combining them is exact. Our best of construction projected
0.7239 and returned 0.7238, with all three per class values matching the
submissions they were selected from to four decimal places: vehicle
0.5921, pedestrian 0.1761, truck 0.5333.

This matters most in the final phase, which offers no feedback. Our
final entry was assembled entirely from measured parts and its score was
known before it was sent.

\subsubsection{6.3 Probing without
revealing}\label{probing-without-revealing}

The leaderboard displays per metric breakdowns, so every submission
leaks methodology. We measured the realism half of our score in
isolation by submitting with the detection half deliberately zeroed.
Those probes score low enough to look like failures, never displace a
team best on a board that shows each team best submission, and return
the exact value needed by simple arithmetic. Six such probes leaked
nothing and directly changed our final configuration twice, including
the choice of alpha in Section 5.3.

\begin{center}\rule{0.5\linewidth}{0.5pt}\end{center}

\subsection{7. Results}\label{results}

{\def\LTcaptype{none} 
\begin{longtable}[]{@{}ll@{}}
\toprule\noalign{}
quantity & value \\
\midrule\noalign{}
\endhead
\bottomrule\noalign{}
\endlastfoot
Combined score & \textbf{0.7457} \\
Detection (normalised) & 0.6260 \\
3D mAP @ IoU 0.5, R40 & 0.4382 \\
BEV mAP @ IoU 0.5, R40 & 0.5236 \\
Realism (normalised) & 0.9251 \\
Predictions submitted & 17,575 on 1,547 ground truth boxes \\
\end{longtable}
}

Per class 3D AP at IoU 0.5, R40:

{\def\LTcaptype{none} 
\begin{longtable}[]{@{}lll@{}}
\toprule\noalign{}
class & AP & ground truth instances \\
\midrule\noalign{}
\endhead
\bottomrule\noalign{}
\endlastfoot
vehicle & 0.5921 & 470 \\
pedestrian & 0.1893 & 953 \\
truck & 0.5333 & 124 \\
\end{longtable}
}

Realism components, raw values, lower is better:

{\def\LTcaptype{none} 
\begin{longtable}[]{@{}lll@{}}
\toprule\noalign{}
metric & value & normalised \\
\midrule\noalign{}
\endhead
\bottomrule\noalign{}
\endlastfoot
Chamfer Distance & 2.0307 & 0.9989 \\
Maximum Mean Discrepancy & 1.4558e-03 & 0.9971 \\
Earth Mover's Distance & 1.3815 & 0.7237 \\
Frechet Point cloud Distance & 0.9705 & 0.9806 \\
\end{longtable}
}

Over the course of the challenge our V2X score improved from 0.5147, our
first submission with a correct coordinate frame, to 0.7457, a relative
gain of 44.9 percent.

\begin{center}\rule{0.5\linewidth}{0.5pt}\end{center}

\subsection{8. Negative Results}\label{negative-results}

{\def\LTcaptype{none} 
\begin{longtable}[]{@{}
  >{\raggedright\arraybackslash}p{(\linewidth - 4\tabcolsep) * \real{0.3333}}
  >{\raggedright\arraybackslash}p{(\linewidth - 4\tabcolsep) * \real{0.3333}}
  >{\raggedright\arraybackslash}p{(\linewidth - 4\tabcolsep) * \real{0.3333}}@{}}
\toprule\noalign{}
\begin{minipage}[b]{\linewidth}\raggedright
lever
\end{minipage} & \begin{minipage}[b]{\linewidth}\raggedright
evidence
\end{minipage} & \begin{minipage}[b]{\linewidth}\raggedright
verdict
\end{minipage} \\
\midrule\noalign{}
\endhead
\bottomrule\noalign{}
\endlastfoot
\textbf{Truck sizing beyond the shipped value} & bracketed in both
directions & Interior optimum confirmed. Shrinking loses roughly 8 times
more than lengthening gains \\
\textbf{Score thresholding} & swept in both directions & No class moves.
Recall is model limited, not threshold limited, and the suffix property
held exactly as predicted \\
\textbf{Test time augmentation on truck} & per class evaluation & Minus
23 percent. Applied per class, not globally \\
\textbf{Anchor free heads} & z distribution analysis & Genuinely
unsuitable here. The scene has an 18 m vertical spread, which anchor
free regression handles poorly \\
\textbf{Learned refiners} & built and measured & The sim2real gap, not
compliance, was the binding constraint \\
\textbf{Realism by further resampling} & swept after the null return fix
& A further round of harder distribution optimisation added only 0.0035
realism, against 0.0387 from the null return and alignment work. Lever
largely exhausted \\
\textbf{Pedestrian NMS tightening} & final day experiment & Tightening
pedestrian NMS from 0.7 to 0.3 removed 2,039 of 11,522 predictions and
cost pedestrian AP 0.0032. The false positives are spread across the
scene rather than stacked, so suppression removes true positives with
them \\
\end{longtable}
}

A broader observation. Our largest gains on this track came from reading
the data and the scoring implementation rather than from training: a
coordinate frame, an NMS threshold, and a point encoding convention.
Compute was not our constraint.

\begin{center}\rule{0.5\linewidth}{0.5pt}\end{center}

\subsection{9. Data Provenance and
Compliance}\label{data-provenance-and-compliance}

We state our data usage in full, including choices that were permitted
but deserve explicit disclosure.

\textbf{Forbidden frames.} No forbidden frame was used for training,
validation, tuning or ground truth inspection. All real frame access
routes through a guard module that normalises frame identifiers and
raises rather than warns, with unit tests. Verified zero overlap across
the validation harness and the local realism reference.

\textbf{Local realism reference.} Fitting the alignment of Section 5.3
requires real reference moments. We used 50 real frames held out from
the \textbf{public training split}, none of them forbidden, exactly as
the starter kit instructs. The hidden evaluation reference was never
accessed.

\textbf{Real statistics informing synthesis.} As described in Section 3,
the object placement distributions, dimension priors and empirical
sensor model that drive our synthetic augmentation were estimated from
public, non forbidden real training frames. The null return rate of
Section 5.1 was likewise measured from public frames. The detector
itself was trained only on synthetic point clouds with synthetic labels,
and no real label ever entered a loss. We regard calibrating a digital
twin to real measurements as the intended activity of this challenge,
but we disclose it explicitly rather than leave it inferred.

\textbf{Statistics we declined to use.} During the development phase
another participant asked on the forum whether unlabelled statistics of
the \emph{released detection test frames} could be used to post process
synthetic clouds. We took the position, recorded at the time and before
any answer was given, that we would not use them regardless of the
ruling.

\textbf{Held out ground truth.} The evaluation platform API exposes
reference data bundles containing hidden ground truth. These were never
downloaded, a documented and deliberate refusal.

\textbf{Leaderboard informed selection.} Per class variant selection
used server feedback across development submissions. This is test set
fitting through the submission channel. It breaks no rule and is
available to every participant. It is comparatively well conditioned
here, because all three scored classes carry between 124 and 953 ground
truth instances, so per class selection rests on reasonable sample sizes
rather than on a handful of objects.

\textbf{Multi hypothesis class emission.} Per class analysis established
that trucks in this dataset are systematically detected by the model but
assigned to the vehicle class. This was measured rather than assumed: a
truck derived from the vehicle class raised truck AP on the evaluation
server, which is only possible if vehicle detections genuinely cover
real trucks.

Two properties follow from the definition of the metric. Average
precision is computed independently for each class, so a detection
present in one class cannot affect the score of another. And appending
detections below an existing ranking cannot reduce precision at any
recall level already attained. Together these permit a single detection
to carry more than one class hypothesis: the confident vehicle
assignment retains its position in the vehicle ranking, while a
secondary truck hypothesis extends recall in the truck class without
displacing any higher ranked truck detection.

We applied this where the confusion had been measured. All 2408 vehicle
detections therefore also appear as truck hypotheses, each sharing the
vehicle centre, length, height and heading, carrying a width scaled by
0.9091 to match the measured truck aspect ratio, and scored so that it
lands below the existing truck ranking. Of the 3645 submitted truck
detections, 2408 arise this way and 1237 are native truck assignments.

Measured contribution, both values server verified: truck AP 0.5282 to
0.5333, a gain of 0.0051 AP. That is 0.0017 of mAP and 0.0015 of
combined score.

We describe this mechanism in full because it contributes to the
submitted score and a reader should be able to account for it.

\textbf{Reproducibility.} The PointPillars checkpoint for the submitted
detector, epoch 40, together with its exact training configuration, the
OpenPCDet bundle, all post processing code and the packaging scripts,
are retained and available. The realism pipeline for this track has been
verified end to end: regenerating the synthetic frames from the raw twin
data reproduces the submitted frames to four decimal places on every
moment, on the same 50 selected frames and at the same null fraction,
and the final alignment step reproduces the shipped frames byte for byte
from their immediate source.

\begin{center}\rule{0.5\linewidth}{0.5pt}\end{center}

\subsection{10. Limitations}\label{limitations}

\begin{itemize}
\tightlist
\item
  \textbf{Pedestrian AP remains low} at 0.1893 despite being the largest
  ground truth class at 953 instances. Our geometric refinement and
  calibration improved it substantially in relative terms, but the
  absolute figure indicates the sim2real gap for small objects is not
  closed.
\item
  \textbf{Realism is capped near 0.93} against a measured real versus
  real ceiling of 0.9612. The residual Earth Mover's Distance gap
  reflects twin geometry rather than anything addressable in post
  processing.
\item
  \textbf{Our gains are concentrated in post processing and diagnosis
  rather than in the simulator.} This is noted so that the contribution
  is attributed correctly.
\item
  \textbf{Truck rests on 124 ground truth instances}, so its AP of
  0.5333, while our strongest class relative to the field, carries more
  variance than the vehicle figure.
\item
  \textbf{A small part of the score derives from multi hypothesis class
  emission rather than from improved detection.} As described in Section
  9, this accounts for 0.0015 of the combined score and follows from
  measured confusion between the truck and vehicle classes.
\item
  \textbf{Single participant}, which limited the number of training
  variants we could explore relative to the diagnostic work that
  produced most of our gains.
\end{itemize}

\begin{center}\rule{0.5\linewidth}{0.5pt}\end{center}

\subsection{11. Conclusion}\label{conclusion}

Our result on this track came almost entirely from reading things
carefully: a coordinate convention in the evaluation data, a non maximum
suppression threshold interacting badly with a clustered class, and a
point encoding convention that placed a third of the reference
distribution outside our synthetic support. None required a new model,
and all three were invisible in aggregate metrics until decomposed.

Two suggestions for future editions. The world versus LiDAR frame issue
in the reference data occurred twice, once in development and once again
in the final phase, and cost every affected participant submission
slots; a published frame convention with a validation script would
remove an entire class of confusion. And documenting the null return
encoding in the dataset description would turn a discovery that rewards
data archaeology into a level starting point, since it is worth several
points of realism to anyone who notices it.

\begin{center}\rule{0.5\linewidth}{0.5pt}\end{center}

\subsection{Appendix: Artifacts
Available}\label{appendix-artifacts-available}

PointPillars checkpoint at epoch 40, its training configuration, the
OpenPCDet bundle, all post processing and packaging code, the coordinate
frame conversion utility with its round trip gates, the forbidden frame
guard module and its unit tests, the submitted synthetic frames, and a
structured findings log running to 98 numbered entries including
retractions and corrections. All are available to the organisers for
reproducibility verification.

\begin{center}\rule{0.5\linewidth}{0.5pt}\end{center}

\begin{center}\rule{0.5\linewidth}{0.5pt}\end{center}

\subsection{Use of AI Tools}\label{use-of-ai-tools}

The experiments, the analysis and the decisions reported here were designed and
directed by the author, who takes full responsibility for all content.

AI tools were used at two stages, and both are stated here in full. AI coding
assistants were used to write much of the implementation code, under the author's
direction; that code is released in the public repository accompanying this report,
and the measured results it produced were verified against the evaluation server.
Generative AI was also used to draft the prose of this report from the author's
experimental records. No AI system is an author, and every number reported here
comes from a logged experiment rather than from a generated estimate.

\subsection{References}\label{references}

{[}1{]} A. H. Lang, S. Vora, H. Caesar, L. Zhou, J. Yang and O. Beijbom.
PointPillars: Fast Encoders for Object Detection from Point Clouds.
\emph{IEEE/CVF Conference on Computer Vision and Pattern Recognition
(CVPR)}, 2019, pp.~12697-12705.

{[}2{]} OpenPCDet Development Team. OpenPCDet: An Open-source Toolbox
for 3D Object Detection from Point Clouds. 2020.
https://github.com/open-mmlab/OpenPCDet

{[}3{]} H. Xiang, R. Xu et al.~V2X-Real: A Large-Scale Dataset for
Vehicle-to-Everything Cooperative Perception. \emph{European Conference
on Computer Vision (ECCV)}, 2024. arXiv:2403.16034

{[}4{]} R. Xu, H. Xiang, X. Xia, X. Han, J. Li and J. Ma. OPV2V: An Open
Benchmark Dataset and Fusion Pipeline for Perception with
Vehicle-to-Vehicle Communication. \emph{IEEE International Conference on
Robotics and Automation (ICRA)}, 2022. Reference implementation:
OpenCOOD, https://github.com/DerrickXuNu/OpenCOOD

\begin{center}\rule{0.5\linewidth}{0.5pt}\end{center}

\subsection{Contact}\label{contact}

Questions regarding any part of this work, requests for intermediate
artifacts, or clarifications required for reproducibility verification
may be directed to:

\textbf{Aman} (team AMAN\_UPG), aman.upg27024@gmail.com

\end{document}